\documentclass[letterpaper]{article} % DO NOT CHANGE THIS
\usepackage{aaai25}  % DO NOT CHANGE THIS
\usepackage{times}  % DO NOT CHANGE THIS
\usepackage{helvet}  % DO NOT CHANGE THIS
\usepackage{courier}  % DO NOT CHANGE THIS
\usepackage[hyphens]{url}  % DO NOT CHANGE THIS
\usepackage{graphicx} % DO NOT CHANGE THIS
\usepackage{natbib}  % DO NOT CHANGE THIS AND DO NOT ADD ANY OPTIONS TO IT
\usepackage{caption} % DO NOT CHANGE THIS AND DO NOT ADD ANY OPTIONS TO IT
\usepackage[ruled,vlined,linesnumbered]{algorithm2e}
\SetCommentSty{emph}
\usepackage{newfloat}
\usepackage{listings}
\DeclareCaptionStyle{ruled}{labelfont=normalfont,labelsep=colon,strut=off} % DO NOT CHANGE THIS
\usepackage{multirow}
\usepackage{graphicx} % for \graphicspath
\usepackage{caption}
\usepackage{subcaption}
\usepackage{url}
\usepackage{xcolor}
\usepackage{amssymb,amsmath,comment,cite}
\usepackage{amsthm}
\usepackage{amsfonts}
\usepackage[capitalize]{cleveref}
\usepackage{booktabs}
\Crefname{equation}{Eq.}{Eqs.}
\Crefname{figure}{Fig.}{Figs.}
\Crefname{tabular}{Tab.}{Tabs.}
\newtheorem{theorem}{Theorem}
\newtheorem{lemma}{Lemma}
\newtheorem{definition}{Definition}
\newcommand{\method}{MASS}
\newcommand{\repolink}{https://github.com/JingtianYan/MASS-AAAI}

\title{Multi-Agent Motion Planning For Differential Drive Robots Through Stationary State Search}
\author {
    Jingtian Yan, %\textsuperscript{\rm 1},
    Jiaoyang Li%\textsuperscript{\rm 1}
}
\affiliations {
    Carnegie Mellon University\\
    jingtianyan@cmu.edu, jiaoyangli@cmu.edu
}

\usepackage{bibentry}
\begin{document}

\maketitle

\begin{abstract}
Multi-Agent Motion Planning (MAMP) finds various applications in fields such as traffic management, airport operations, and warehouse automation.
In many of these environments, differential drive robots are commonly used.
These robots have a kinodynamic model that allows only in-place rotation and movement along their current orientation, subject to speed and acceleration limits.
However, existing Multi-Agent Path Finding (MAPF)-based methods often use simplified models for robot kinodynamics, which limits their practicality and realism.
In this paper, we introduce a three-level framework called \method \ to address these challenges. 
\method \ combines MAPF-based methods with our proposed stationary state search planner to generate high-quality kinodynamically-feasible plans.
We further extend \method \ using an adaptive window mechanism to address the lifelong MAMP problem.
Empirically, we tested our methods on the single-shot grid map domain and the lifelong warehouse domain.
Our method shows up to 400\% improvements in terms of throughput compared to existing methods.
\end{abstract}
\section{Introduction}
We study the Multi-Agent Motion Planning (MAMP) problem which aims to find collision-free kinodynamically feasible paths for a team of agents in a fully observable environment while minimizing their arrival time.
This problem finds various real-world applications, including traffic management~\cite{Ho2019traffic}, airport operations~\cite{li2019departure}, and warehouse automation~\cite{kou2019multi}. 
Differential drive robots are widely used in many of these environments. These robots, often navigating on a grid map, can move forward along their orientation with bounded velocity and acceleration.
They can only change their orientation through in-place rotation when at zero speed.
Although much work has been done to address the MAMP problem, existing methods often either fail to account for the orientations of robots or overlook continuous dynamic constraints.

\begin{figure}[!t]
\centering
    \includegraphics[width=\linewidth]{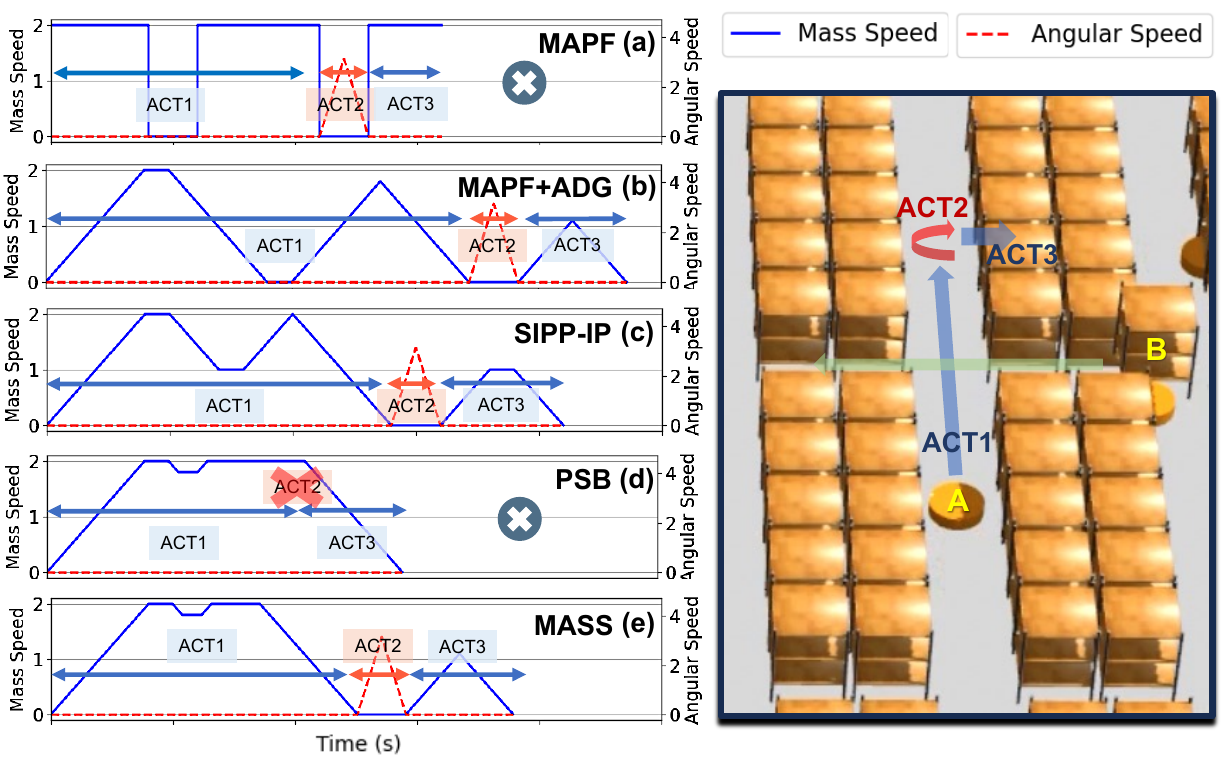}
    \caption{Speed profile of agent A (blue line: linear velocity, red line: angular velocity) generated by (a) MAPF, (b) MAPF+ADG, (c) SIPP-IP, (d) PSB, and (e) MASS. Agent A first moves upward (ACT1) while adjusting its speed to avoid collisions with agent B, performs an in-place rotation (ACT2), and then moves to the right (ACT3).}
\label{fig:mapf_intro}
\end{figure}

Multi-Agent Path Finding (MAPF)~\cite{Stern2019benchmark} methods are a promising solution that scales to hundreds of agents.
However, they assume instantaneous movement and infinite acceleration capabilities, resulting in plans that are unrealistic for real-world execution (see \cref{fig:mapf_intro} (a) for an example).
To apply MAPF methods to MAMP, ADG~\cite{honig2019warehouse} post-processes the speed profiles of the MAPF plan to meet kinodynamic constraints while maintaining the passing orders of agents at each location. 
However, as shown in~\cref{fig:mapf_intro} (b), such methods can lead to long execution time as the initial MAPF plan (\cref{fig:mapf_intro} (a)) overlooks kinodynamic constraints. 
SIPP-IP~\cite{ali2023safe} extends MAPF methods to MAMP by searching with a fixed number of predefined actions with discretized speeds and accelerations.
However, as shown in~\cref{fig:mapf_intro} (c), due to its discretized nature, the limited action choices can lead to long execution time or even failures in solving certain cases.
Moreover, to account for the different choices of speeds and accelerations, SIPP-IP explores a high-dimensional state space which compromises its efficiency.
The recent work PSB~\cite{yan2024PSB} avoids such discretization by combining search-based and optimization-based methods, but it does not consider the orientations of agents, making it hard to apply to differential drive robots, as shown in~\cref{fig:mapf_intro} (d).

In this work, we introduce \textbf{\underline{MA}}PF-\textbf{\underline{S}}SIPP-\textbf{\underline{S}}PS (\method), a framework to address the MAMP problem for differential drive robots.
\method \ uses a key observation that the plan for these robots always alternates between rotation and movement.
Thus, the state at the transition between two actions (finish a movement to start rotation or the reverse) is critically important.
Since those states must have zero speed, we refer to them as \emph{stationary states}.
Instead of searching at high-dimensional state space that models various speeds of robots, we focus our search on these stationary states and actions that connect them.
In \method, we use a MAPF-based planner at Level 1 to resolve collisions between agents.
This level imposes temporal constraints on Level 2 and calls it to get a plan for each agent.
We propose Stationary Safe Interval Path Planning (SSIPP) at Level 2 to search for a single-agent kinodynamically feasible plan.
Compared to standard SIPP~\cite{phillips2011sipp}, SSIPP uses stationary node expansion to find neighboring stationary states and the actions needed to reach them. 
At Level 3, an optimization-based speed profile solver (SPS) is used to determine the speed profiles for these actions.

Our main contributions include:
1. We propose a framework called \method, a three-level MAMP planner capable of finding collision-free plans for a large group of differential drive robots.
2. We evaluate \method\ on the standard MAPF benchmark, showing significant improvement in terms of success rate, especially for large-scale maps.
3. We extend \method\ to address the lifelong MAMP problem where agents are assigned new goals after they reach their current ones.
We evaluate \method\ in a high-fidelity automated warehouse simulator (shown in \Cref{fig:exp_sim}). 
\method\ shows up to 400\% improvement in terms of solution cost compared to a MAPF planner with a post-processing framework.
\section{Background}
In this section, we begin with a review of MAPF algorithms.
After that, we go through the related work in MAMP.

\subsubsection{MAPF Algorithms}
MAPF methods have achieved remarkable progress in finding discrete collision-free paths for hundreds of agents.
Most state-of-the-art MAPF methods, such as Conflict-Based Search (CBS)~\cite{sharon2015conflict, andreychuk2022multi} and Priority-Based Search (PBS)~\cite{ma2019searching}, use a bi-level structure.
At the high level, they resolve collisions among agents by introducing temporal obstacles into low-level single-agent solvers. These solvers then plan paths for individual agents trying to avoid those temporal obstacles.
% These methods find collision-free temporal paths for multiple agents by planning a path for each agent while considering other agents as dynamic obstacles. When planning for one agent, the method assumes the paths of other agents as temporary obstacles and tries to avoid collisions with them during the low-level planning process.
In our experiments, we test MASS with two MAPF algorithms, PP and PBS. In Priority Planning (PP)~\cite{erdmann1987multiple}, the planner begins by assigning a total priority ordering to all agents
% where the agent with lower priority must avoid collisions with the agent with higher priority., 
requiring lower-priority agents to avoid collisions with higher-priority ones. 
Then, the PP plans paths for each agent from high priority to low priority.
During this process, agents treat the path from higher priority agents as temporal obstacles. Priority-Based Search (PBS)~\cite{ma2019searching} searches for a good priority ordering that prevents collisions among agents. 
PBS explores a binary Priority Tree (PT) in a depth-first manner, where each PT node contains a set of partial priority orderings and corresponding paths.
The root node starts with no priority orderings.
When a collision between agents $a_i$ and $a_j$ is detected, the PT is expanded by creating two child nodes, each with an additional priority ordering $i \prec j$ or $j \prec i$, indicating $a_i$ has higher or lower priority than $a_j$. 
In each child node, PBS uses a low-level planner to replan the paths based on the updated priority orderings. The search terminates when a PT node with collision-free paths is found.

\subsubsection{MAMP Algorithms}
The first category of MAMP methods directly extends single-agent motion planners~\cite{vcap2013multi}. These methods combine the state space of individual agents into a collective joint space to perform single-agent motion planning. 
Since the dimension of this space increases exponentially in the number of agents, planning within the joint state space of agents presents scalability challenges.
Another category of methods uses the MAPF methods to solve the MAMP problem. 
Some of them use discrete paths from MAPF planners to generate trajectories that meet kinodynamic constraints \cite{honig2016multi, zhang2021temporal, honig2019warehouse}. 
For instance, the Action Dependency Graph (ADG)~\cite{honig2019warehouse} generates speed profiles for each agent based on discrete MAPF plans.
It post-processes the speed profiles of the MAPF plan to meet kinodynamic constraints while maintaining the passing orders of agents at each location by encoding the action-precedence relationships.
% During execution, an action is only allowed for execution once all its action-precedence conditions are met.
The solution quality of these methods highly relies on the discrete paths from MAPF planners.
However, as discussed in~\cite{varambally2022mapf}, since the MAPF planners use an inaccurate kinodynamic model, their solution quality is often limited.
Some other methods extend MAPF methods to consider robot kinodynamics during planning.
These methods typically discretize the action space and use a graph-search-based method~\cite{solis2021representation, cohen2019optimal, ali2023safe}.
However, with their discretized nature, they consider a limited number of actions and thus fail to capture the full range of possible actions that agents could exhibit.
Moreover, they also face scalability challenges as they search in a high-dimensional state space.
To avoid such discretization, PSB~\cite{yan2024PSB} combines search-based and optimization-based methods to produce solutions with smooth speed profiles.
% However, PSB cannot optimize paths where the spatial movements and the temporal movements are coupled, making it inefficient in handling in-place rotation.
However, PSB cannot handle the in-place rotation of agents, making it hard to apply to differential drive robots.
The existing work closest to ours is the extended abstract by \citet{kou2019multi}. Instead of searching the high-dimensional state space, they suggest performing an $A^*$ search over the states with speeds of zero and show promising preliminary results.
Our \method\ is inspired by this idea.

\section{Problem Formulation} \label{sec:formulation}

\begin{figure*}[!t]
\centering
    \includegraphics[width=.9\linewidth]{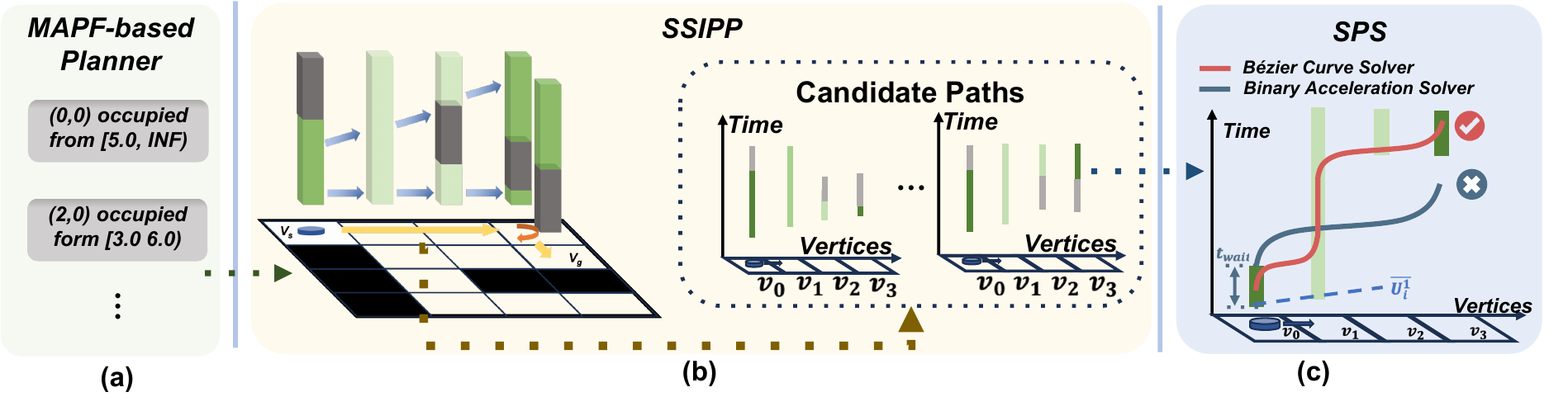}
    \caption{System overview. In (b), the green strips are safe intervals, the dark green strips are stationary safe intervals, and the gray boxes are temporal obstacles given by Level 1.}
\label{fig:system_overview}
\end{figure*}

We define our MAMP problem with differential-drive agents as the MAMP$_{D}$ problem on an undirected graph $G = (V, E)$ and a set of $M$ agents $\mathcal{R} = \{a_1, ..., a_M\}$.
% We define our MAMP$_{D}$ problem by a undirected graph $G = (V, E)$ and a set of $M$ agents $\mathcal{R} = \{a_1, ..., a_M\}$.
We adopt the grid model from the MAPF$_R$ problem~\cite{walker2018extended} and represent $G$ as a four-neighbor grid map.
Vertices in \( V \) represent grid cells in the map, with their locations the same as the center of each cell and shapes equal to the cell size.
An edge $(v_i, v_j) \in E$ corresponds to possible transitions between $v_i$ and $v_j$.
We use a differential drive robot model with a specific shape.
When at a vertex, an agent can have a discretized orientation $\theta \in \Theta$.
We define the \emph{state} of an agent as a collection of its vertex, orientation, and speed at a specific time.
Each agent $a_m$ initiates its movement from a specified \textit{start (vertex)} $v_{s_m} \in V$ and start orientation $\theta_s \in \Theta$, where $\Theta$ is a finite set of possible orientations.
Each agent $a_m$ has a single designated \textit{goal (vertex)} $v_{g_m} \in V$.
All agents start simultaneously and remain at their respective goals after they finish.
An agent can perform one of the following actions at each vertex with the action time $T_{m}$ being the time it takes to finish this action:
\begin{definition}
(Rotate) A rotate($\theta_i$, $\theta_j$) lets an agent change its orientation from $\theta_i \in \Theta$ to $\theta_j \in \Theta$ on its current vertex.
This action begins and ends with the agent at zero speed and follows a predefined angular velocity profile. 
The action time of $rotate(\theta_i,\theta_j)$ is no greater than the sum of the action time of $rotate(\theta_i,\theta_k)$ and $rotate(\theta_k,\theta_j)$ for all $ \theta_k$.
\end{definition}
\begin{definition}
(Move) A move($v_i$, $v_j$) lets an agent move forward in its current orientation from $v_i$ to $v_j$ along a straight line segment $\phi_{i,j}$, which may include one or more vertices.
This action begins and ends with the agent at zero speed and follows a \textit{speed profile} $\ell_{i,j}(t \mid \phi_{i,j})$, denoted as the distance traveled by an agent as a function of time $t$ along a given line segment $\phi_{i,j}$.
For agent $a_m$, the speed profile of its move action is constrained by the following dynamic constraints:
\begin{equation}
% \small
    \underline{U_m^k} \le \frac{{d^k \ell_{i,j}(t \mid \phi_{i,j})}}{{d t^k}} \le \overline{U_m^k}, \forall k \in \{1, 2\} \label{eq:kinodymaic_constraints}
\end{equation}
\begin{equation}
% \small
    \left.\frac{d \ell_{i,j}(t \mid \phi_{i,j})}{d t}\right\vert_{t=0, T_{m}} = \underline{U_m^1}, \label{eq:init}
\end{equation}%
where $\underline{U_m^k}$ and $\overline{U_m^k}$ represent the lower and upper bounds of speed (when $k=1$) and acceleration (when $k=2$), respectively, with the minimum speed being $\underline{U_m^1}=0$.
The planner needs to determine a speed profile (including $T_m$) for each move action.
We define a move action as dynamically feasible if its speed profile satisfies these constraints.
\end{definition}

A \textit{timed action} represents an action that starts at a specific time.
A plan $p_m$ of $a_m$ is a set of timed actions that move $a_m$ from its start to its goal.
An agent \textit{reaches} a vertex iff the geometric centers of the vertex and the agent overlap.
We define that an agent \textit{occupies} a vertex \textit{v} if its shape overlaps the shape of \textit{v}.
A \textit{collision} occurs if two agents occupy the same \textit{v} and at overlapping time intervals.
We use \textit{arrival time} to indicate the time needed for $a_m$ to reach $v_{g_m}$.
Our task is to generate plans for all agents so that no collisions happen while minimizing the sum of their arrival time.

\subsubsection{Lifelong MAMP$_{D}$}
Compared to the single-shot MAMP formulation, in the lifelong MAMP model, there are two main differences:
(D1) Each agent receives new goals assigned by an external task assigner during execution and must visit these assigned goals sequentially.
(D2) Agents are not required to stay at their goals, instead, they must perform one of the three additional actions at each goal, namely attaching themselves to a shelf, detaching themselves from a shelf, or waiting at a station.
Our task is to maximize the throughput (= average number of reached goal vertices in a certain time duration).
\section{\method} \label{sec:method}
In this section, we begin with a system overview of our proposed method, \method, followed by the specifics of the SSIPP used in Level 2.
Next, we introduce a partial stationary expansion mechanism to improve its scalability.
Then, we present the formulation for speed profile optimization and two example solvers.
Finally, we discuss the techniques used to extend \method\ to the lifelong MAMP scenario.

\subsection{System Overview}
\noindent
\textbf{\textit{MAPF-based Planner}}
At Level 1, we borrow the MAPF-based planner to resolve collisions between agents.
Our framework is compatible with any MAPF solver employing a bi-level structure as discussed in related work.
Empirically, we use PP and PBS as the Level-1 planner.

\noindent
\textbf{\textit{Stationary SIPP (SSIPP)}}
The task of Level 2 is to find a plan for an agent with minimum arrival time while avoiding temporal obstacles (e.g., paths of higher priority agents from PP or PBS) given by Level 1. As shown in~\cref{fig:system_overview} (b), we first build a safe interval table $\mathcal{T}$ based on those temporal obstacles.
This table associated each vertex of $G$ with a set of \emph{safe intervals}, which are time intervals not occupied by the temporal obstacles.
Then, Level 2 performs an SSIPP search on $\mathcal{T}$ to find the neighbor stationary states along with the actions that lead to them, where the speed profile of these actions is found by Level 3.
We use a partial stationary expansion (PE) mechanism to further improve its efficiency.

\noindent
\textbf{\textit{Speed Profile Solver (SPS)}}
The task of Level 3 is to find a speed profile that travels within safe intervals given by Level 2, satisfies dynamic constraints, and achieves optimal action time.
We introduce two solvers in this section: the Binary Acceleration Solver, an incomplete but fast method, and the B\'ezier-Curve Solver, a complete but slow method.

\subsection{Stationary SIPP (SSIPP)}
Given a safe interval table $\mathcal{T}$, the task of Level 2 is to find a collision-free plan for agent $a_m$ while minimizing its arrival time.
In our problem, agents move in continuous time with continuous dynamics, leading to an infinite number of possible states at each vertex.
To address this, Level 2 employs SSIPP, which performs an A* search on $\mathcal{T}$ to avoid directly searching through such states.
Compared to standard SIPP~\cite{phillips2011sipp}, SSIPP uses stationary node expansion to find stationary states and dynamically feasible actions connecting them.
In the rest of this section, we omit subscript $m$ for simplicity. 

\subsubsection{SSIPP Node}
The search node of SSIPP is defined as $n = \{v, \theta, a, [lb, ub)\}$. $v \in V$ and $\theta \in \Theta$ are the vertex and orientation of the agent. $a$ is the previous action that leads the agent to node $n$. 
$[lb, ub)$ is a stationary safe interval, a specific safe interval in which the agent can maintain a stationary state at $v$.

\subsubsection{Main Algorithm}
Algorithm~\ref{alg:SSIPP} shows the pseudo-code of SSIPP.
We begin by initializing the root node with start vertex $v_s$, start orientation $\theta_s$, and the first interval at $v_s$ in $\mathcal{T}$ [Line \ref{alg:line:init}].
Then we push the root node to an open list OPEN [Line \ref{alg:line:insert}].
The arrival time of the best plan $p^*$ is initially set to infinity [Line \ref{alg:line:init_arrival}].
We define the $g$-value of a node $n$ as its $lb$-value, its $h$-value as the minimum time to move from its vertex to the goal, and its $f$-value as the sum of its $g$- and $h$-values, which is a lower bound on the arrival time of any plan that goes through $n$ (i.e. stops at $n$ within its time interval).
At each iteration, we select the node $n$ with the smallest $f$-value from OPEN [Line \ref{alg:line:select}].
If the $f$-value of $n$ is bigger than the arrival time of $p^*$, it indicates that $p^*$ is the optimal plan.
We terminate the search [Line~\ref{alg:ssipp:break}] and return $p^*$ [Line~\ref{alg:line:return}].
If $n$ is at the goal with infinite $n.ub$, we check if its $g$-value is smaller than the arrival time of $p^*$. If true, we update $p^*$ by backtracking all ancestor nodes of $n$ [Line~\ref{alg:ssipp:reach}-\ref{alg:ssipp:getplan}]. In either case, we proceed to the next iteration.
For all other nodes, we use stationary node expansion to generate new neighbor nodes and push them to OPEN [Line~\ref{alg:ssipp:expansion}].
This search proceeds until the optimal plan is found or OPEN is empty.

\begin{algorithm}[t]
\small
\caption{Stationary SIPP (SSIPP)}
\label{alg:SSIPP}
\DontPrintSemicolon
\SetKwFunction{FOPEN}{pushToOPEN}

\KwIn{start vertex $v_s$, start orientation $\theta_s$, goal vertex $v_g$, safe interval table $\mathcal{T}$}
$root\_n \gets (v_s, \theta_s, \textit{none}, {\color{blue} \emptyset}, \mathcal{T}[v_s][0])$\; \label{alg:line:init}
\FOPEN($root\_n$)\; \label{alg:line:insert}
\textit{p*.arrival\_time}$\gets \infty$ \; \label{alg:line:init_arrival}
\While{OPEN $\neq \emptyset$}{
    $n \gets$ OPEN.pop()\; \label{alg:line:select}
    \lIf{$n.f \ge \text{p*.arrival\_time}$}{%
        \Return $p*$ \label{alg:ssipp:break}
    }
    \If{$n.v = v_g$ \textbf{and} $n.ub = \infty$ \label{alg:ssipp:reach}}{ 
        \lIf{$n.g < \textit{p*.arrival\_time}$}{$p* \gets getPlan(n)$ \label{alg:ssipp:getplan}}
        \textbf{continue}
    }
    $\texttt{{\color{blue} (partial)}StationaryNodeExpansion}(n)$\; \label{alg:ssipp:expansion}
    % {\color{blue} $\mathrm{partialNodeExpansion(n)}$} \;
}
\Return ``No solution found''\label{alg:line:return}

\SetKwFunction{FMove}{createNodeByMove}
\SetKwFunction{FMain}{stationNodeExpansion}
\SetKwProg{Fn}{Function}{}{}
\Fn{\FMain{n}}{
        \If(\tcp*[f]{rotate expansion}){$n.a \neq rotate$}{
            % Operations for partial node expansion if the previous action was 'move'
            $\{n'_0,...,n'_j\} \gets \texttt{rotateExpansion}(n)$\; \label{alg:rotate_exp}
            \FOPEN{$n'_0,...,n'_j$} \label{alg:insert_rotate}
        
        }
        \If(\tcp*[f]{move expansion}){$n.a \neq move$ \label{alg:line:if_not_move}} {
            $\mathbb{S} \gets \texttt{getMoveIntervals}(n)$\; \label{alg:getmove}
            \For{$[lb, ub) \in \mathbb{S}$}{
            \FMove{$n, [lb, ub)$}\label{alg:call_gen_move_func}
            }
        }
}
\SetKwProg{Fn}{Function}{}{}
    \Fn{\FMove{$n, [lb, ub)$}}{
    % Node Re-expansion: Find Forward Nodes
    $(\phi, S) \gets \texttt{backTrack}([n.lb, n.ub), [lb, ub))$\; \label{alg:partial:getpath_interval}
    $\ell(t) \gets \texttt{getSpeedProfile}(\phi, S)$\; \label{alg:partial:solve}
    \If{$\ell(t) \neq null$}{
    $n' \gets (vertex([lb, ub)), n.\theta, \textit{move},{\color{blue} \emptyset},$ $[n.lb+$actionTime($\ell$)$, ub))$\;\label{alg:partial:new_node}
    \FOPEN{$n'$} \label{alg:partial:insert_move_to_open}
    }
}
\end{algorithm}

\subsubsection{Stationary Node Expansion}
At each stationary state, we let the agent perform an action different from its previous action; otherwise, two identical actions can be combined into one.
Accordingly, stationary node expansion includes two types: move expansion and rotate expansion. 
Rotate expansion finds all neighbor nodes reachable through rotation, while move expansion does the same for movement.

\noindent
\emph{Rotate Expansion}:
Since the orientation is discretized, during rotation expansion, we apply all the possible predefined rotation speed profiles (i.e., rotate $90^\circ$, $-90^\circ$, and $180^\circ$) to generate the neighbor nodes of a given node $n = \{v, \theta, a, [lb, ub)\}$.
Specifically, we create a new neighbor node $n' = \{v, \theta', rotate, [lb', ub)\}$ for each possible orientation $\theta' \in \Theta$, with $lb'$ being the sum of $lb$ and the rotation time. $n'$ is discarded if $lb' \ge ub$.

\begin{figure}[!t]
\centering    
\includegraphics[width=\linewidth]{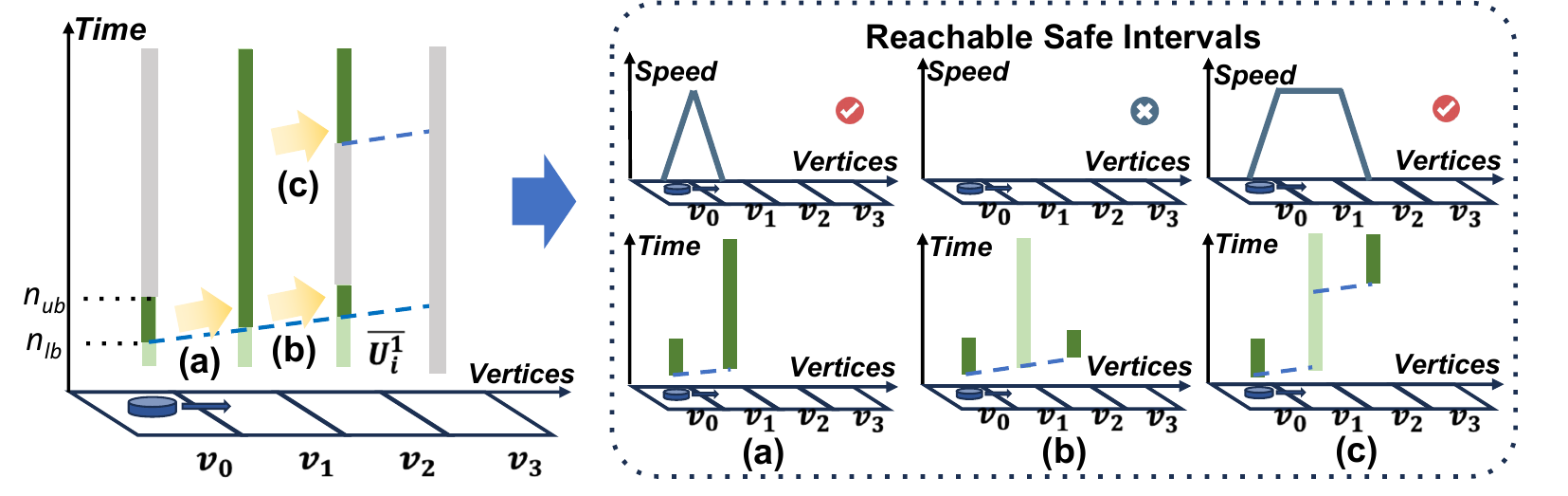}
    \caption{Illustration of the safe interval search process.}
\label{fig:interval_search}
\end{figure}

\noindent
\emph{Move Expansion}:
During move expansion, we first find all safe intervals in $\mathcal{T}$ at all vertices that may be reached through a move action from the current node, referred to as \emph{reachable intervals}.
Then, for each reachable interval, we treat it as a stationary safe interval and use Level 3 to find a speed profile to reach it.
If a speed profile is found, we generate a new SSIPP node for this safe interval.

Concretely, if node $n$ is the root node or its previous action is a rotate action, we first call \texttt{getMoveIntervals} which uses a breadth-first search on safe intervals along the node's orientation to find all reachable intervals [\cref{alg:line:if_not_move,alg:getmove}].
We use an example in~\cref{fig:interval_search} to illustrate this process.
We begin by initializing the root interval using the interval of the current node and push it to a queue.
During each iteration, we pop an interval from the queue and expand it by assuming the agent moves one vertex forward.
In our case, we first expand the interval at $v_0$ denoted as $[lb_0, ub_0)$.
As the agent moves from $v_{0}$ to $v_{1}$, a new interval $[lb_0 + t_{min}, \infty)$ is generated at $v_{1}$, where $t_{min}$ is the minimum time required for this movement.
Following~\cite{yan2024PSB}, since dynamic constraints are considered in Level 3, we can use relaxed dynamic constraints to expedite this expansion process without compromising the guarantee of completeness. 
Specifically, we estimate $t_{min}$ as the time the agent takes to move at maximum speed.
For safe intervals at $v_1$ with a lower bound smaller than $ub_0$, which are the intervals that can be directly reached from $[lb_0, ub_0)$, we treat their overlap with $[lb_0 + t_{min}, \infty)$ as stationary safe intervals.
We get the safe intervals shown in ~\cref{fig:interval_search} (a) in our example and push it to a reachable interval set $\mathbb{S}$.
In the next iteration, we continue to expand the safe intervals at $v_1$.
This search process proceeds recursively until no stationary safe intervals can be found.

For each reachable interval $[lb, ub) \in \mathbb{S}$, we call \texttt{createNodeByMove} to generate an SSIPP node [\cref{alg:call_gen_move_func}].
We backtrack to get all safe intervals $S=\{[lb_0, ub_0), ..., [lb, ub)\}$ along with its associated line segment $\phi$ [Line~\ref{alg:partial:getpath_interval}].
Then, we call Level 3 to find a speed profile $\ell(t)$ based on them [Line~\ref{alg:partial:solve}].
Using $\ell(t)$ found by Level 3, we generate a new node at the associated vertex of $[lb, ub)$ and push it to OPEN [Line~\ref{alg:partial:new_node}-\ref{alg:partial:insert_move_to_open}].

\subsubsection{Duplicate Detection}
A duplicate detection mechanism is used to eliminate redundant nodes during the search. Before inserting a node $n$ into the open list, we check whether a node with the same vertex, orientation, and upper bound already exists in the open list or has been visited. If a duplicate node $n'$ is found, we compare the lower bounds of $n$ and $n'$, and retain the node with the smaller lower bound.
We prove that this duplicate detection mechanism does not affect the completeness or the optimality of \method.

\begin{algorithm}[t]
\small
\caption{Partial Stationary Expansion} \label{alg:PE-SIPP}
\DontPrintSemicolon
\SetKwFunction{FPartial}{partialStationaryNodeExpansion}
\SetKwProg{Fn}{Function}{}{}
\Fn{\FPartial{n}}{
    \If(\tcp*[f]{expand node $n$ for the first time}){$n.\mathcal{F} = \emptyset$}{
        \If{$n.a \neq rotate$}{
            % Operations for partial node expansion if the previous action was 'move'
            $\{n'_0,...,n'_j\} \gets \texttt{rotationExpansion}(n)$\; \label{alg:partial:rotate_exp}
            \FOPEN($n'_0,...,n'_j$) \label{alg:partial:insert_rotate}
        }
        \If{$n.a \neq move$} {
            $n.\mathcal{F} \gets$\texttt{getMoveIntervals}($n$) \label{alg:partial:getmove}\;
            Sort intervals in $n.\mathcal{F}$ by their $p$-values
        }
    }
    \If(\tcp*[f]{generate one child node}){$n.\mathcal{F} \neq \emptyset$}{
        \FMove{$n, n.\mathcal{F}.pop()$}\; \label{alg:partial:createnode}
    }
    \If(\tcp*[f]{reinsert node $n$}){$n.\mathcal{F} \neq \emptyset$}{
        $n.h \gets n.\mathcal{F}.top().p-n.g$\; \label{alg:partial:update_heuristic}
        \FOPEN{$n$}\label{alg:partial:reinsert}
    }
}
\end{algorithm}

\begin{theorem}
    [Completeness and optimality of SSIPP]
    SSIPP is complete and returns the optimal solution if one exists when Level 3 is complete and optimal. Please refer to the Appendix for detailed proof.\label{theorem:completeness_SSIPP}
\end{theorem}
% \begin{proof}
    % Given Level 3 is complete, we can explore all safe intervals, ensuring that a solution will be found if it exists.
    % Meanwhile, since the $f$-value of an SSIPP node is the lower bound of its arrival time from this node, the smallest $f$-value in OPEN, $f^*$, bounds the arrival times of all paths in OPEN. When $f^* \geq \textit{p*.arrival\_time}$, no paths in OPEN can arrive earlier than optimal path $p^*$.
    % Please refer to the Appendix for detailed proof.
% \end{proof}

\subsection{Partial Stationary Expansion (PE)}
During move expansion, we need to find the speed profiles for all reachable safe intervals.
This branching factor can be very high, especially in large maps. 
To tackle this, we use a partial stationary expansion mechanism extended from~\cite{goldenberg2014enhanced}.
\subsubsection{PE Node}
This node extends the SSIPP node by $n = \{v, \theta, a, \mathcal{F}, [lb, ub)\}$, where 
\textit{Reachable interval list} $\mathcal{F}$ is a list that contains all the reachable safe intervals, denoted as $\{[lb_0, ub_0), ...\}$.
The intervals in $\mathcal{F}$ are sorted in ascending order of their $p$-value (= $lb$ plus $h$-value at its associated vertex), which is an underestimate of the arrival time through this interval.

\subsubsection{Partial Stationary Node Expansion}
In partial stationary node expansion, instead of finding the speed profiles for all reachable intervals at once, we only generate the node based on the reachable interval that is most promising.
As shown in~\cref{alg:PE-SIPP}, if $n.\mathcal{F}$ is empty and its previous action is \textit{move}, we do rotate expansion to retrieve its neighbor nodes [Line~\ref{alg:partial:rotate_exp}-\ref{alg:partial:insert_rotate}].
Otherwise, instead of performing move expansion, we only retrieve all the reachable intervals for $n.\mathcal{F}$ [Line~\ref{alg:partial:getmove}].
In case $n.\mathcal{F}$ is not empty, we pop the reachable interval with the smallest $p$-value in $n.\mathcal{F}$ and generate a neighbor node based on it [Line~\ref{alg:partial:createnode}].
Finally, if $n.\mathcal{F}$ remains non-empty, we update the heuristic value of $n$ using the smallest $p$-value in $n.\mathcal{F}$ and reinsert $n$ into OPEN [Line~\ref{alg:partial:update_heuristic}-\ref{alg:partial:reinsert}].

\begin{theorem}
    [Completeness and optimality of SSIPP with PE]
    The partial expansion mechanism preserves the completeness and optimality of SSIPP.
Detailed proof is provided in the Appendix. \label{theorem:completeness_PE}
\end{theorem}
% \begin{proof}
% While PE generates neighbor nodes selectively, it still considers all reachable intervals, ensuring completeness of \method\ unchanged.
% At the same time, we re-insert the node with updated $f$-value back to OPEN.
% The updated $f$-value is a lower bound of the $f$-values of all the ungenerated neighbor nodes.
% Similar to~\cref{theorem:completeness_SSIPP}, we can prove the optimality of \method\ is preserved when using partial stationary expansion.
% Detailed proof is provided in the Appendix.
% \end{proof}

\subsection{Speed Profile Solver (SPS)}
Given the line segment $\phi_{i,j}$ and safe intervals $S$ from Level 2, SPS aims to find a speed profile $\ell_{i,j}(t)$ with the shortest action time that satisfies both the dynamic constraints shown in \cref{eq:kinodymaic_constraints,eq:init} and temporal constraints introduced by $S$ (i.e., the agent remains within the safe interval while passing a vertex).
This section introduces two SPS as examples.
Notably, \method \ is adaptable to other solvers, as long as it meets the specified constraints.

\subsubsection{Binary Acceleration Solver (BAS)}
We adopt BAS from~\cite{kou2019multi}.
This solver assumes that the agent begins by waiting at the first vertex $v_i$ of the line segment $\phi_{i,j}$ for a duration of $t_{wait}$.
Then, it moves with its maximum acceleration until reaching its maximum speed, moves at this speed for a duration of $t_{move}$, and finally decelerates with its maximum deceleration to stop at $v_j$.
However, when the length of $\phi_{i,j}$ is small, the speed profile forms a triangle shape, where the agent accelerates to a lower peak speed and then decelerates to stop at $v_j$.
$t_{move}$ can be computed based on the length of $\phi_{i,j}$. 
Our task is to get the $t_{wait}$ that minimizes action time while ensuring that $\ell_{i,j}(t)$ satisfies the temporal constraints
introduced by $S$. This problem can be formulated as a Linear Programming (LP) problem.
We borrow \Cref{fig:system_overview} (c) as a counterexample to show BAS is incomplete.
In this case, a valid speed profile exists where the agent waits at $v_1$. However, BAS fails to find this solution because it only decelerates upon reaching the goal.

\subsubsection{B\'ezier-curve Solver (BCS)}
We borrow BCS from~\cite{yan2024PSB}.
BCS models the speed profile using a scaled B\'ezier curve, which can approximate any continuous function within its feasible range with sufficient control points.
BCS encodes the temporal and dynamic constraints as an LP problem and then uses binary search to determine the optimal action time by solving this LP problem recursively.
As shown in the paper, given any $\epsilon$, BCP can find a speed profile $\epsilon$-close to the optimal solution with a sufficient number of control points if one exists and returns failure otherwise.

\subsection{Lifelong MAMP$_D$}
% Motivation
% Given the difference between the models outlined.
% First introduce the difficulty of directly applying RHCR to our method
In this section, we extend \method \ to address the lifelong MAMP$_D$ problem.
Many works have been done to extend the single-shot MAPF problem to the lifelong scenario.
In this work, we adapt the state-of-the-art method Rolling-Horizon Collision Resolution (RHCR)~\cite{li2021lifelong} to \method.
RHCR decomposes the lifelong MAPF problem into a sequence of windowed MAPF instances.
Specifically, it plans collision-free paths for $t_w$ timesteps and replans paths once every $t_h$ timesteps ($t_w \geq t_h$).
However, in \method, the actions can have arbitrary action time. 
As a result, we can no longer determine a fixed replanning window size $t_w$ that guarantees that all agents have just completed their actions and arrived at vertices at time $t_w$.
In this work, we incorporate an adaptive window mechanism that apply different window sizes for different agents.

\begin{algorithm}[t]
\small
\caption{Pseudocode for Windowed-SSIPP}\label{alg:window-SIPP}
\DontPrintSemicolon
\KwIn{earliest start time $t_{e}$, goal list $\mathcal{G}$ and safe interval table $\mathcal{T}$}
$root\_n \gets (v_s, \theta_s, \textit{none}, \emptyset, \mathfrak{g}=\mathcal{G}[0], [t_{e}, \mathcal{T}[v_s][0].ub))$\; \label{alg:win_sipp:rootnode}
\FOPEN(\(root\_n\))\; \label{alg:win_sipp:insert_rootnode}
$n^*.f_{win} \leftarrow \infty$\;
\While{OPEN $\neq \emptyset$}{
    \(n \gets\) OPEN.pop()\; \label{alg:win_sipp:pop_n}
    \lIf{\(n.f_{win} > n^*.f_{win}\)}{
        $getPlan(n*)$
    } \label{alg:win_sipp:find_solution}
    \lIf{\(n.t_{ub} = \infty\) \textbf{and} \(n^*.f_{win} > n.f_{win}\)}{
        \(n^* \gets n\)
    } \label{alg:win_sipp:update_optimal}
    \If{\(n.v = n.\mathfrak{g}\) \textbf{and} \(n.lb+\textit{actionTime}(\mathcal{G}[n.l].a)<n.ub\)}{
        \(n' \gets (n.v, n.\theta, n.\mathfrak{g}.a, \emptyset, \mathcal{G}.next(\mathfrak{g}), \) \([n.lb+\textit{actionTime}(\mathcal{G}[n.l].a), n.ub))\)\; \label{alg:win_sipp:gen_new}
        \FOPEN(\(n'\))\; \label{alg:win_sipp:insert_new}
        % \textbf{continue}\; \label{alg:win_sipp:continue}
    }
    \If{\(n.t_{lb} < t_w\)}{
        \FPartial{n} \label{alg:win_sipp:partial_expansion}
        % Assuming Partial Node Expansion is defined as Window-F.
    }
}
\Return ``No solution found''
\end{algorithm}

\begin{figure}[!t]
\centering
    \includegraphics[width=.78\linewidth]{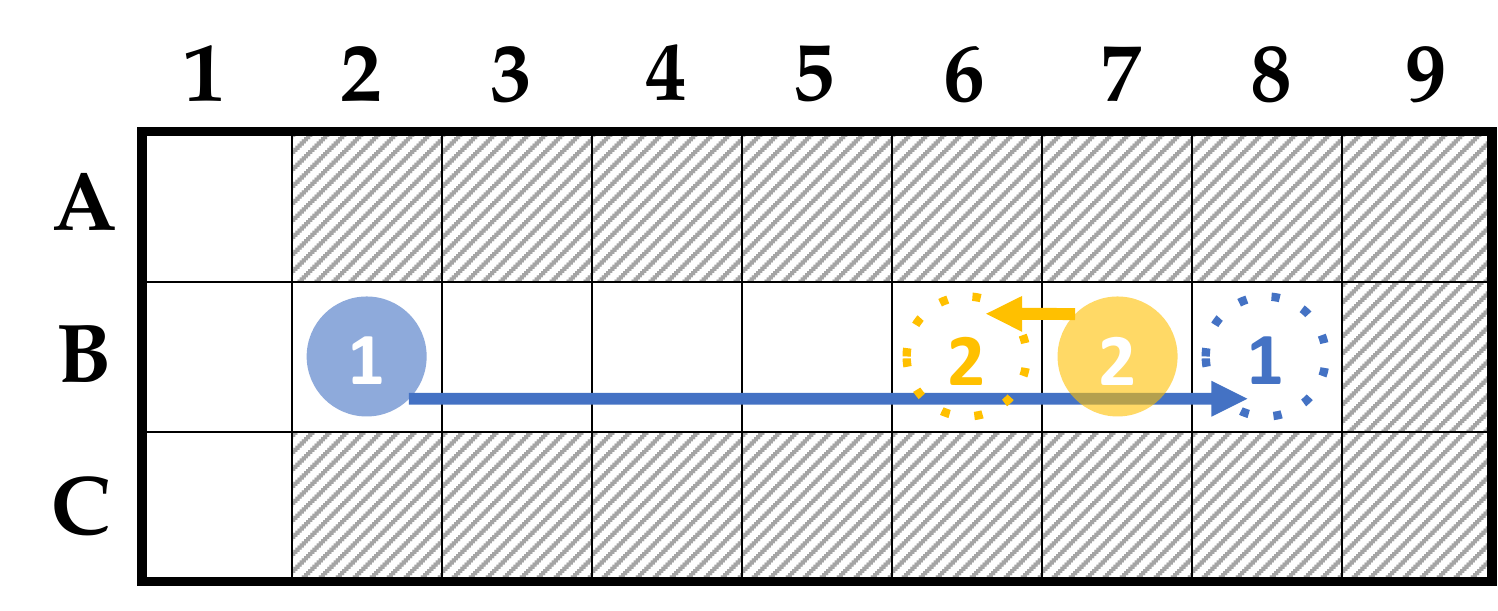}
    \caption{Limitation of directly applying RHCR to \method. Solid circles are the current locations of agents and dashed circles are the expected start location for the next episode.}
\label{fig:sipp_window}
\end{figure}

\subsubsection{Adaptive Window}
% We borrow the concept of replanning window $t_h$ and collision-free horizon $t_{w}$ from RHCR.
Similar to RHCR, we trigger replanning every $t_h$ time duration to plan for the next episode. However, since planning for a fixed episode length $t_w$ is not feasible, $t_w$ serves only as the minimum size of the replanning window. 
We define actions that start before and finish after $t$ as $t$-\textit{crossing actions}. 
Within each episode, the adaptive window plan of each agent consists of the $t_w$-\textit{crossing action}, along with any preceding actions.
During planning, we ensure that adaptive window plans are collision-free between agents.
Since adaptive window plans can have arbitrary window sizes, the earliest start time for each agent at the next episode may vary.
As agents must complete ongoing actions, for each agent, we determine its start location and its earliest start time for the next episode based on the final vertex and the completion time of its $t_h$-\textit{crossing action} from the current episode.
However, as shown in~\cref{fig:sipp_window}, this approach introduces a new issue:
Consider agents $a_1$ and $a_2$, where both their adaptive window plans at the current episode consist only of a $t_w$-\textit{crossing action} (shown as the blue and yellow arrows in~\cref{fig:sipp_window}). We assume they are also $t_h$-\textit{crossing actions}. These plans are initially collision-free because the $t_w$-\textit{crossing action} of $a_2$ occupies B6 only until it completes.
However, in the next episode, when $a_2$ starts from B6, no feasible plan exists to avoid a collision.
To resolve this, we always append a waiting action at the end of each plan. In this case, using the same example, the adaptive window plan of $a_1$ will collide with $a_2$.

The windowed mechanism, which aims to find the node with the minimum $f$-value, introduces another issue: when using the standard $f$-value, agents may prefer to wait within $t_w$ rather than move forward.
For example, when $a_1$ in~\cref{fig:sipp_window} moves from B2 to B8 using an underestimated heuristic, the $f$-value at B2 could be smaller than its neighbors, causing it to wait at B2.
To address this, we introduce the $f_{win}$-value, defined as $n.f_{win} = \max(t_w, n.g) + n.h$. This penalizes agents for stopping prematurely before $t_w$.

\begin{figure*}[!t]
\centering
    \includegraphics[width=.9\linewidth]{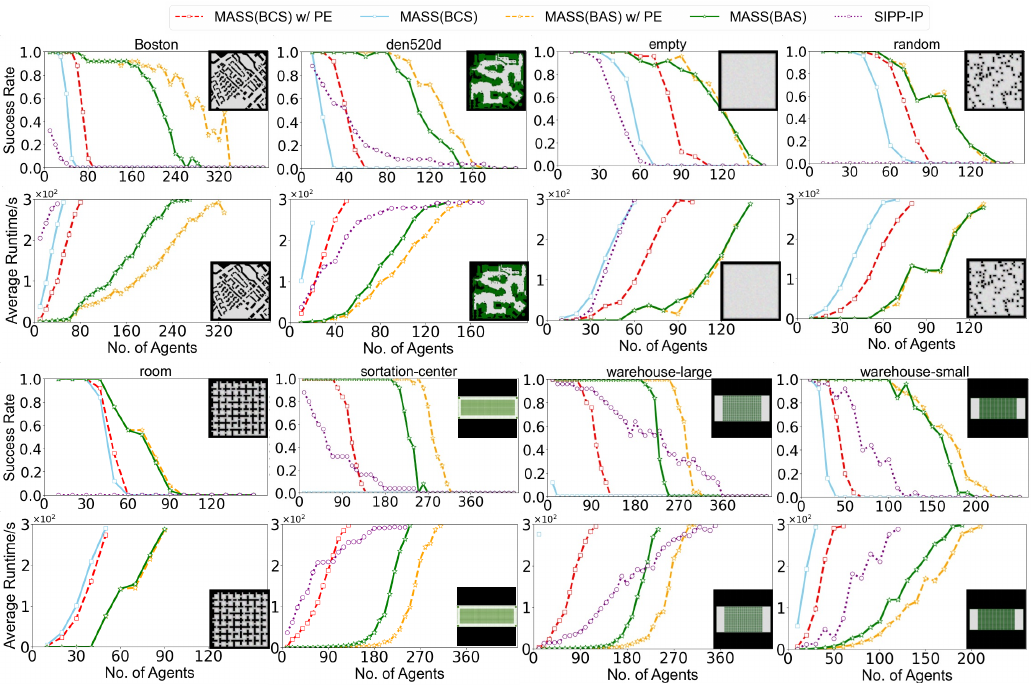}
    \caption{Success rate and average runtime across all maps.
     The success rate is the ratio of solved instances to all instances.}
\label{fig:exp_mapf}
\end{figure*}

\begin{figure}[!t]
\centering
\includegraphics[width=.96\linewidth]{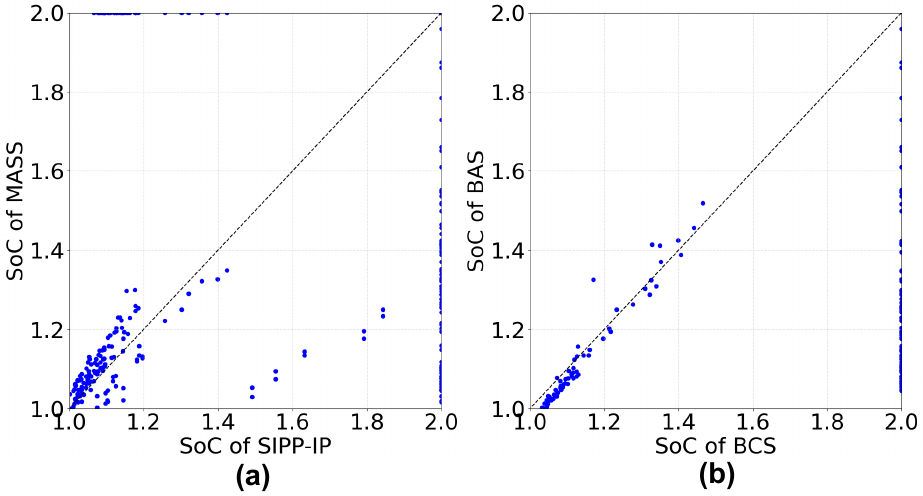}
    \caption{
    Relative SoC is the ratio of the total arrival time to the sum of individual agent arrival times without considering collisions. We use points at 2.0 to indicate unsolved instances.
    Figure (a) shows the relative SoC of SIPP-IP compared to MASS (both BAS and BCS). Figure (b) shows the relative SoC between MASS(BAS) and MASS(BCS).}
\label{fig:exp_mapf_sol_cost}
\end{figure}

% 1. Introduce how to determine multiple goals
\subsubsection{Main Method}
In every episode, for each agent $a_m$, we first update the start location $v_{s_m}$, the earliest start time $t_{e_m}$, and goal list $\mathcal{G}$.
Here, the goal list stores the goal vertices in the order they should be visited.
% We determine $v_{s_m}$ and $t_{e_m}$ based on the last vertex and the time it was reached in the crossing action from the previous episode.
Then, we call \method\ to find the plans for this episode.
Here, Level 1 and Level 3 can be applied without any modification.
As shown \cref{alg:window-SIPP}, we begin the search process of Level 2 by initializing the root node with the first goal in $\mathcal{G}$ and pushing it to OPEN [Line~\ref{alg:win_sipp:rootnode}-\ref{alg:win_sipp:insert_rootnode}].
During each iteration, we find the node $n$ with the smallest $f_{win}$-value and remove it from OPEN [Line~\ref{alg:win_sipp:pop_n}].
If the $f_{win}$-value of $n$ is larger than the $f_{win}$-value of the optimal node $n*$ found so far, we terminate the search and backtrack to return the plan [Line~\ref{alg:win_sipp:find_solution}].
Otherwise, in case $n.ub$ is infinite and $n$ has a smaller $f_{win}$-value than $n*$, we update $n*$ using $n$ [Line~\ref{alg:win_sipp:update_optimal}].
% search optimal end node (node that can stay)
If $n.v$ equals the goal vertex $n.\mathfrak{g}$ that $n$ is trying to reach, we generate a new node $n'$ that performs the required action $n.\mathfrak{g}.a$ at vertex $n.\mathfrak{g}$ and set its goal using the next element in $\mathcal{G}$ [\cref{alg:win_sipp:gen_new,alg:win_sipp:insert_new}].
Finally, if the lower bound of $n$ is smaller than $t_w$, we do partial stationary expansion [Line~\ref{alg:win_sipp:partial_expansion}].
If \method\ is unable to find a solution within the given cutoff time, we reuse the adaptive window plan from the previous episode and continue its execution.

\section{Empirical Evaluation}
We implemented both our and baseline methods in C++. We conducted all experiments on an Ubuntu 20.04 machine equipped with an AMD 3990x processor and 188 GB of memory.
Our code was executed using a single core for all computations.
The source code for our method is publicly accessible at \repolink.

\subsection{Single-Shot MAMP$_D$}
In this experiment, we use PBS as Level 1 and both BAS and BCS as Level 3 . We denote the resulting two variants as \method(BAS) and \method(BCS). 
They are further combined with the partial expansion mechanism, denoted as \method(BAS) w/ PE and \method(BCS) w/ PE.
We compare these methods with a straightforward extension of SIPP-IP~\cite{ali2023safe}. 
SIPP-IP is a state-of-the-art single-agent safe interval path planner designed to accommodate kinodynamic constraints and temporal obstacles, making it a suitable representation of motion-primitive-based methods.
To adapt SIPP-IP for multi-agent scenarios, we replaced Level 2 and Level 3 in \method \ with the SIPP-IP.

\begin{table}
\small
% \scriptsize % Or \footnotesize
% \setlength{\tabcolsep}{3pt}
% \renewcommand{\arraystretch}{0.9}
    \centering
    \begin{tabular}{c|c|c|c}
        \toprule
        \multicolumn{2}{c|}{Runtime (s)} & MASS(BCS) w/ PE & MASS(BAS) w/ PE \\
        \midrule
        \multirow{4}{*}{\rotatebox{90}{10 Agents}} 
        & Total & 34.17\(\pm\)24.3\% & 0.17\(\pm\)0.4\% \\
        & PBS & 0.00\(\pm\)0.0\% & 0.03\(\pm\)0.0\% \\
        & SIPP & 0.02\(\pm\)0.0\% & 0.13\(\pm\)0.3\% \\
        & SPS & 34.16\(\pm\)24.3\% & 0.01\(\pm\)0.0\% \\
        \midrule
        \multirow{4}{*}{\rotatebox{90}{150 Agents}} 
        & Total & nan & 170.50\(\pm\)114.3\% \\
        & PBS & nan & 20.49\(\pm\)18.0\% \\
        & SIPP & nan & 131.12\(\pm\)88.2\% \\
        & SPS & nan & 18.88\(\pm\)13.5\%\\
        \bottomrule
    \end{tabular}
    \caption{Runtime breakdown of MASS on the \texttt{warehouse-small} map in seconds. PBS, SIPP, and SPS are the runtime of each component.}
    \label{tab:runtime}
\end{table}

\subsubsection{Simulation Setup}
We evaluated all methods on four-neighbor grid maps, including \texttt{empty} (empty-32-32, size: 32$\times$32), \texttt{random} (random-32-32-10, size: 32$\times$32), \texttt{room} (room-64-64-8, size: 64$\times$64), \texttt{den520d} (den520d, size: 256$\times$257), \texttt{Boston} (Boston\_0\_256, size: 256$\times$256), \texttt{warehouse-small} (warehouse-10-20-10-2-1, size: $161\times63$), \texttt{warehouse-large} (warehouse-20-40-10-2-2, size: $340\times164$) from the MovingAI benchmark \cite{Stern2019benchmark}, and the \texttt{sortation-center} map (size: $500\times140$) from the LMAPF Competition~\cite{chan2024league}.
For each map, we conducted experiments with a progressive increment in the number of agents, using the 25 ``random scenarios'' from the benchmark set.
The agents were modeled as cycles with a diameter equal to the length of the grid cell.
All agents adhered to the same kinodynamic constraints, where the speed is bounded by the range of $[0, 2]\; cell/s$, while the acceleration is confined to $[-0.5, 0.5]\; cell/s^2$.

\subsubsection{Comparison}
As shown in~\cref{fig:exp_mapf}, PE improves the success rate for \method(BCS) on all maps.
For \method(BAS), it improves the success rate in large-scale maps, while maintaining comparable results on small-scale maps. 
This improvement is primarily because PE shows advantages when Level 3 is time-consuming (e.g., using BCS) or the branching factor during move expansion is high (e.g., on large maps).
Compared to BCS, BAS demonstrates its advantage in terms of success rate.
As shown in~\cref{fig:exp_mapf_sol_cost} (b), despite BCS being a complete and optimal method, BAS achieves a similar solution cost.
We hypothesize this is due to the scalability limitations of BCS, as it can only handle less congested cases where BAS also provides near-optimal solutions.
SIPP-IP has a low success rate in obstacle-rich maps due to its limited action choice.
At the same time, as shown in~\cref{fig:exp_mapf_sol_cost} (a), it shows worse solution quality than \method \ in certain cases.

\subsubsection{Runtime}
As shown in~\cref{tab:runtime}, we include the runtime details of \method\ with different SPS on \texttt{warehouse-10-20-10-2-1} map.
The primary runtime bottleneck for \method(BAS) lies in Level 2 (SIPP) for both the 10-agent and 150-agent cases. In contrast, \method(BCS) is limited by Level 3 (SPS) in 10-agent scenarios and fails to scale to scenarios with a larger number of agents.
% Compared to the methods without a partial expansion mechanism, the methods with it show better total runtime.
% Meanwhile, methods using BAS as SPS show much lower runtime, indicating we should use more scalable SPS in resource-constrained settings.

\begin{figure}[!t]
\centering
\includegraphics[width=1.0\linewidth]{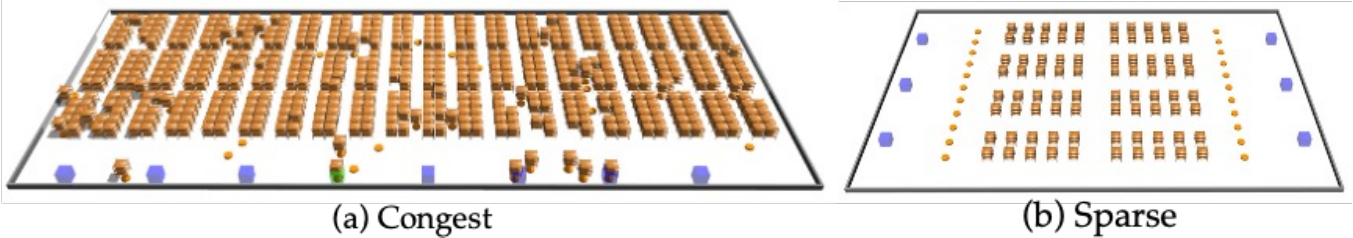}
    \caption{Simulation setup. (a) contains 8 stations, 600 shelves, and 50 agents (based on the iRobot Create 2).
    (b) contains 6 stations, 80 shelves, and 22 agents.}
\label{fig:exp_sim}
\end{figure}

\begin{table}[t]
\small
\centering
\begin{tabular}{c|c|c|c|c}
\toprule
% \multirow{2}{*}{Env} & \multirow{2}{*}{w} &  \multicolumn{3}{c|}{Throughput} \\ \cline{3-5}
%  & & \multicolumn{1}{c|}{\textbf{MASS(BAS)}} & \multicolumn{1}{c|}{\textbf{MASS(BCS)}} & \multicolumn{1}{c|}{\textbf{SIPP w/ ADG}} \\ \hline 
Env & $t_w$ &  \multicolumn{1}{c|}{\textbf{MASS(BAS)}} & \multicolumn{1}{c|}{\textbf{MASS(BCS)}} & \multicolumn{1}{c}{\textbf{PP w/ ADG}} \\ 
\midrule
 \multirow{4}{*}{\rotatebox{90}{Sparse}} 
 & 20 s & \textbf{0.262$\pm$2.9\%} & 0.215$\pm$6.4\% & 0.142$\pm$8.4\% \\ \cmidrule{2-5}
 & 25 s & \textbf{0.264$\pm$4.3\%} & 0.215$\pm$7.0\% & 0.152$\pm$8.3\% \\ \cmidrule{2-5}
 & 30 s & \textbf{0.259$\pm$4.5\%} & 0.215$\pm$1.9\% & 0.149$\pm$6.8\% \\ \cmidrule{2-5}
 & 40 s & \textbf{0.259$\pm$3.0\%} & 0.218$\pm$3.7\% & 0.148$\pm$7.1\% \\ 
\midrule
 \multirow{4}{*}{\rotatebox{90}{Congest}} 
 & 20 s & \textbf{0.373$\pm$1.4\%} & 0.294$\pm$4.4\% & 0.065$\pm$2.5\% \\ \cmidrule{2-5}
 & 25 s & \textbf{0.380$\pm$3.4\%} & 0.300$\pm$3.4\% & 0.075$\pm$4.0\% \\ \cmidrule{2-5}
 & 30 s & \textbf{0.371$\pm$4.1\%} & 0.294$\pm$13.4\% & 0.090$\pm$3.4\% \\ \cmidrule{2-5}
 & 40 s & \textbf{0.372$\pm$6.7\%} & 0.300$\pm$6.4\% & 0.095$\pm$2.4\% \\ 
 \bottomrule
\end{tabular}
\caption{Throughput in \textit{Congest} and \textit{Sparse}.}
\label{table:exp_lifelong}
\end{table}

\subsection{Lifelong MAMP$_D$}
% \subsubsection{Baseline Algorithms}
In this experiment, we use PP with random start as Level 1 for \method(BAS) and \method(BCS), incorporating the partial expansion mechanism.
We use PP w/ ADG~\cite{varambally2022mapf} to represent methods that combine MAPF with a robust execution framework.
PP w/ ADG uses RHCR to decompose the lifelong MAPF problem into windowed MAPF instances, uses PP with SIPP for planning in each window, and uses 
ADG to execute the plans.

\subsubsection{Simulation Setup}
We borrow the simulation setup from~\cite{honig2019warehouse} to simulate a Kiva warehouse on the \textit{Congest} map with 50 agents and \textit{Sparse} map with 22 agents, using Amazon's HARMONIES simulator, as shown in~\cref{fig:exp_sim}.
Each agent has a speed limit from $[0, 2]\; m/s$ and an acceleration limit from $[-0.5, 0.5]\; m/s$.
We run each method for 1,000 simulation time seconds and average the results over 7 runs.

\subsubsection{Comparison}
We evaluate solution quality using the \textit{throughput} (=average goals reached per second).
As shown in \cref{table:exp_lifelong}, the throughput of \method(BAS) and \method(BCS) are significantly better than PP w/ ADG.
This indicates that incorporating the kinodynamics of agents during the planning process can improve the solution quality.
At the same time, \method(BAS) achieved a slight improvement in throughput compared to \method(BCS). 
This is attributed to the factor that BAS is able to explore more priority orderings within the given time window due to its shorter runtime.

\section{Conclusion}
This paper introduces \method, a three-level multi-agent motion planning framework designed to tackle the MAMP problem for differential drive robots. 
\method \ uses SSIPP to search the stationary state along with actions between them.
We further add a partial stationary expansion mechanism to improve its scalability and extend \method \ to the lifelong MAMP$_D$ domain.
Empirically, \method \ shows significant improvements in both scalability and solution quality compared to existing methods.
% Based on both one-shot and lifelong experiments, it seems that BAS is consistently better than BCS. 

\section*{Acknowledgments}
The research was supported by the National Science Foundation (NSF) under grant number \#2328671 and a gift from Amazon.
The views and conclusions contained in this document are those of the authors and should not be interpreted as representing the official policies, either expressed or implied, of the sponsoring organizations, agencies, or the U.S. government.
\bibliography{aaai25}
% \newpage
% \input{docs/checklist}
\newpage
\section*{Appendix}

% \begin{figure*}[!t]
% \centering
%     \includegraphics[width=.9\linewidth]{imgs/exp/result_mapf_v2.pdf}
%     \caption{Success rate, average cost, and average runtime across all maps.
%      The success rate is the ratio of solved instances to all instances.}
% \label{fig:exp_mapf_all}
% \end{figure*}

\subsection*{Theoretical Proofs}
\begin{lemma}
    Given travel time for $rotate(\theta_i,\theta_j)$ is no greater than the sum of travel time for $rotate(\theta_i,\theta_k)$ and $rotate(\theta_k,\theta_j), \forall \theta_k$, duplicate detection mechanism does not affect the completeness of SSIPP.
\end{lemma}

\newcommand{\duration}{D}

\begin{proof}
    Assume we have two SSIPP nodes $n$ and $n'$ where $n.v = n'.v$, $n.\theta = n'.\theta$, and $n.ub=n'.ub$, if $n.lb<n'.lb$, we can only keep $n$ in the OPEN without losing completeness.
    If $n.a=n'.a$, since both $n$ and $n'$ can perform the same type of node expansion (either rotate expansion or move expansion), we can easily prove that pruning $n'$ does not compromise completeness. 
    
    In the case where $n.a \neq n'.a$, it indicates that $n'$ can generate different child nodes during expansion. 
    Let $n_p$ denote the parent node of $n$, with $n_p.a = n'.a$. For any child node generated by $n'$, $n_p$ can generate a corresponding node at the same vertex and orientation.
    To prove completeness, we show that the child nodes generated by $n_p$ at the same vertex and orientation always have a smaller lower bound than those generated by $n'$. 
    Let $\overline{n}_p$ and $\overline{n}'$ represent the nodes generated at a given vertex and orientation by $n_p$ and $n'$, respectively. We define $\duration(n_1, n_2)$ as the minimum time required to transition from the state in $n_1$ to the state in $n_2$.
    The lower bound for $\overline{n}_p$ can be represented by $\overline{n}_p.lb = n_p.lb + \duration(n_p, \overline{n}_p)$, while $\overline{n}'.lb = n'.lb + \duration(n', \overline{n}')$.
    From definition, we can have:
    \begin{align}
        n.lb = n_p.lb + \duration(n_p, n)\\
        \overline{n}_p.lb = n.lb - \duration(n_p, n) + \duration(n_p, \overline{n}_p) \label{eq:proof:lb_parent}\\
        \duration(n, \overline{n}_p) = \duration(n', \overline{n}') \label{eq:proof:equal}
    \end{align}
    Given $\duration(n_p, \overline{n}_p) \leq \duration(n_p, n) + \duration(n, \overline{n}_p)$, this means moving directly from $n_p$ to $\overline{n}_p$ takes no more time than taking two steps. Combine this with~\cref{eq:proof:lb_parent} we can have:
    \begin{align}
        \overline{n}_p.lb & = & n.lb - \duration(n_p, n) + \duration(n_p, \overline{n}_p)\\
        & \leq & n.lb + \duration(n, \overline{n}_p) \label{eq:proof:inequal}
    \end{align}
    Since $n.lb < n'.lb$, using~\cref{eq:proof:equal,eq:proof:inequal} we can have:
    \begin{align}
        \overline{n}_p.lb & \leq & n.lb + \duration(n, \overline{n}_p) \\
        & < & n'.lb + \duration(n', \overline{n}') \\
        & = & \overline{n}'.lb
    \end{align}
    Thus, we prove that the child nodes generated by $n'$ always have a higher lower bound than those generated by $n_p$ at the same vertex and orientation.
\end{proof}

\noindent\textbf{Theorem~\ref{theorem:completeness_SSIPP}.} (Completeness and optimality of SSIPP). \textit{SSIPP is complete and returns the optimal solution if one exists when Level 3 is complete and optimal. }
\begin{proof}
    We begin by proving the completeness of SSIPP followed by the proof of its optimality.
    Since $\mathcal{T}$ contains a finite number of safe intervals, the search space of SSIPP is finite. Thus, SSIPP can terminate within a finite time if no solution exists. During the stationary node expansion, given that Level 3 is complete, we can explore all reachable safe intervals, ensuring that a solution will be found if it exists.
    We then prove the optimality of SSIPP. 
    Since the $f$-value of an SSIPP node is a lower bound on the arrival times of its corresponding plan, the smallest $f$-value of the SSIPP nodes in OPEN, denoted as $f(n)$, is a lower bound on the arrival times of the corresponding paths of the SSIPP nodes in OPEN. 
    When $f(n) \ge travel\_time(p*)$, no corresponding paths of the SSIPP nodes in OPEN can have shorter arrival times than $p*$.
    SSIPP is complete and optimal.
\end{proof}

\noindent\textbf{Theorem~\ref{theorem:completeness_PE}.}
    (Completeness and optimality of SSIPP with PE).
    \textit{The partial expansion mechanism does not change the completeness and optimality of SSIPP.}
\begin{proof}
    The completeness of SSIPP is maintained because the partial stationary expansion mechanism, while delaying the exploration of some nodes, does not exclude any node from eventual expansion. 
    As the search progresses, all nodes with the potential to lead to a solution are eventually expanded, ensuring that a solution, if one exists, will be found.
    We then prove the optimality is preserved when using partial stationary expansion.
    Even though not all children of a node are expanded immediately, we re-insert their parent node with updated $f$-value back to OPEN.
    The updated $f$-value is an underestimation of those child nodes.
    Thus, we can reuse the proof from~\cref{theorem:completeness_SSIPP}.
    Therefore, partial expansion retains the completeness and optimality in SSIPP.
\end{proof}

% \subsection{Additional Experiment Results}
% \cref{fig:exp_mapf_all} shows some additional results on single-shot MAMP experiments.
% In this figure, we use relative SoC, which is the ratio of the total arrival time to the sum of the optimal arrival time for each agent, to measure the solution quality.
% Similar to the success rate, the average runtime of \method \ shows significant improvement compared to SIPP-IP.

\end{document}